\documentclass{article}
\usepackage{microtype}
\usepackage{graphicx}
\usepackage{booktabs}
\usepackage[hyperfootnotes=false]{hyperref}

\usepackage[accepted]{eiml_icml2026}

\makeatletter
\renewcommand{\Notice@String}{\textit{2nd Workshop on Epistemic Intelligence in Machine Learning
(EIML@ICML 2026), Seoul, South Korea. Copyright 2026 by the author(s).}}
\makeatother

\usepackage{amsmath,amssymb,amsthm,mathtools}
\usepackage{enumitem}
\usepackage{doi}
\usepackage[capitalize,noabbrev]{cleveref}
\usepackage{placeins}
\usepackage{dblfloatfix}
\hypersetup{
  pdfauthor={Akira Okutomi},
  pdfsubject={EIML@ICML 2026 Workshop Camera-Ready}
}

\graphicspath{{figures/}}
\makeatletter
\def\input@path{{Tables/}}
\makeatother

\icmltitlerunning{Stable Miscalibration in Large Language Models}

\begin{document}

\twocolumn[
\icmltitle{Stable Miscalibration in Large Language Models: A Practical View of High-Confidence Errors}

\begin{icmlauthorlist}
  \icmlauthor{Akira Okutomi}{tms}
\end{icmlauthorlist}

\icmlaffiliation{tms}{ToppyMicroServices O\"U, Tallinn, Estonia}

\icmlkeywords{Large Language Models, Uncertainty, Calibration, Abstention, Stability}

\vskip 0.3in
]

\makeatletter
\gdef\icmlcorrespondingauthor@text{Akira Okutomi, Tokyo, Japan}
\makeatother

\printAffiliationsAndNotice{}

\pagestyle{fancy}
\fancyhf{}
\fancyhead[C]{\small\bf Stable Miscalibration in Large Language Models}
\fancyfoot[C]{\thepage}
\renewcommand{\headrulewidth}{1pt}

\begin{abstract}
High-confidence errors in large language models are often treated as evidence of fragile internal inference.
We study a different possibility: stable miscalibration, where a confident wrong answer remains locally stable under small perturbations.
We combine two diagnostics: a label-aware output-level audit score that ranks domains by confidence variation and overconfident mistakes under a forced-answer baseline, and an internal sensitivity probe that measures hidden-state movement.
On a multi-domain binary factual audit set, this audit score tracks where abstention-aware self-critique reduces decision loss, although direct labeled baselines rank the same gain more strongly.
Internally, self-critical prompting consistently reduces hidden-state sensitivity across layers in three open-weight models.
This supports prompt-induced local stabilization rather than a purely output-level abstention pattern, but it does not imply calibration: audit-defined overconfident errors are not clearly more locally sensitive than confidently correct answers, so some high-confidence errors may be stable and miscalibrated rather than simply fragile.
\end{abstract}

\section{Introduction}
\label{sec:intro}

High-confidence hallucinations in large language models (LLMs) are often described as a sign
of internal fragility: if a small prompt or context change can flip the answer, then a
high-confidence error looks like a fragile failure mode
\cite{ji2023survey,manakul2023selfcheckgpt,joo2025consistency}. That picture is plausible,
but it is not the only one. A wrong answer can also be \emph{stable}: the model may stay
near the same internal state under small perturbations and still produce a confident
mistake. We refer to this possibility as \emph{stable miscalibration}.

This distinction matters for uncertainty estimation. If high-confidence errors mainly come
from local fragility, then perturbation sensitivity should separate overconfident mistakes
from correct high-confidence answers. If they instead reflect stable miscalibration, then
high-confidence errors may matter for decisions without being unusually fragile. For
hard-to-interpret LLMs, a full internal explanation is often out of reach, so we focus on
what can be learned from confidence responses and small internal probes.

We therefore ask a narrower but sharper question: when high-confidence errors matter in
practice, what can we infer from the confidence responses and perturbation behavior that we
can actually measure?
High-confidence errors can be viewed as failures of epistemic awareness: the model acts as
if its evidence is sufficient even when the decision remains unreliable.

Our work sits at the intersection of abstention-aware decision policies
\cite{wen2025abstentionSurvey,tayebati2025conformalabstention,an2025teachingabstain},
calibration and overconfidence in LLMs
\cite{guo2017,kadavath2022,xiao2025restorecalibration,wang2026faithfulconfidence}, and
perturbation or consistency probes for factual reliability
\cite{ji2024_internal_states_hallu_risk,manakul2023selfcheckgpt,joo2025consistency}. We make
three claims.
\begin{enumerate}[leftmargin=*]
  \item \textbf{Stable miscalibration is a useful uncertainty hypothesis.} A
    linear--Gaussian closed-loop abstraction suggests that local stability and correctness
    need not coincide.
  \item \textbf{A label-aware audit score ranks where intervention helps.} A domain-level
    score built from policy-level confidence variation and overconfident mistakes under a
    forced-answer baseline correlates with where abstention-aware self-critique reduces
    policy-aware loss. Because the score uses observed correctness, it should be read as a
    labeled audit diagnostic.
  \item \textbf{Prompt-induced local stabilization is visible internally, but it is not a fragility separator.}
    Across three open-weight models, the self-critical prompt lowers layer-wise hidden-state
    sensitivity. However, audit-defined overconfidently wrong items do not show a clear
    sensitivity gap relative to audit-defined confidently correct items.
\end{enumerate}

The main point is not a full mechanistic reduction, but a practical lesson:
uncertainty-related behavior that matters for decisions can sometimes be studied through
confidence responses even when the internal mechanism remains only partly understood.

\section{Related Work}
\label{sec:related}

The closest related work falls into three groups. First, abstention-aware
prediction treats refusal as a decision option rather than a failure mode, and recent LLM
work emphasizes that the usefulness of abstention depends on both calibration and task value
\cite{wen2025abstentionSurvey,tayebati2025conformalabstention,an2025teachingabstain}. Our
policy-aware loss follows this view: the question is not only whether a policy is accurate,
but whether it answers in the right places.

Second, calibration work shows that language models can remain overconfident, and recent
studies question whether verbalized confidence reliably turns into good decisions
\cite{guo2017,kadavath2022,xiao2025restorecalibration,wang2026faithfulconfidence}. We build
on this literature, but our goal is narrower than a new calibration benchmark. We ask
whether a practical score built from observable confidence behavior can rank where
intervention is likely to help.

Third, perturbation and consistency probes use either internal representations or output
changes to study reliability under small input variation
\cite{ji2024_internal_states_hallu_risk,manakul2023selfcheckgpt,joo2025consistency}. We use
both views at once: an output-level score for domain ranking and a layer-wise internal probe
for testing whether overconfident mistakes are actually more locally fragile.

\section{A Practical View of Stable Miscalibration}
\label{sec:operational}

We use a linear--Gaussian closed-loop model only as a simple abstraction~\cite{kalman1960},
not as a literal model of LLM inference. A minimal local version is
\begin{equation}
z_{t+1}=Az_t+b+\eta_t,\qquad \eta_t\sim\mathcal N(0,\Sigma).
\end{equation}
If $\rho(A)<1$, small perturbations are damped and the state converges locally. The fixed
point $z^\star=(I-A)^{-1}b$, however, can still lie in a decision region whose label
disagrees with the truth. Thus local stability is compatible with systematic miscalibration.
This is the picture we call \emph{stable miscalibration}. Here we focus on what this picture
implies for quantities we can measure directly.

The practical takeaway is simple. If stable miscalibration is real, then two observable
signals become important. First, confidence may move substantially when the same item is
examined under nearby decision policies. Second, the reference policy may accumulate a
noticeable mass of confident mistakes even when its behavior is locally stable. Our domain
score is designed to capture exactly those two signals.

For LLMs, we summarize domain-level risk with an observable score built from confidence
movement and an overconfident-wrong (OCW) indicator:
\begin{equation}
\begin{aligned}
H_{\mathrm{proxy}}(d)
&= 
\frac{1}{|D_d|}
\sum_{i \in D_d}
\Big[
\mathrm{Std}_{c \in \{C0,C1,C2\}}(p_{i,c}) \\
&\qquad\qquad
+\;
\lambda\,\mathrm{OCW}(i)
\Big].
\end{aligned}
\label{eq:eiml-hproxy}
\end{equation}
Here C0 is the forced-answer baseline, C1 is the cautious-abstention policy, and C2 is the
self-critical abstention policy (see Section~\ref{sec:setup}). The value $p_{i,c}\in[0,1]$ is the reported
$P(\mathrm{correct})$ for item $i$ under policy $c$, with abstentions mapped to the neutral
value $0.5$. The overconfident-wrong (OCW) indicator is $1$ for an item when C0 is overconfidently wrong: it
answers the item, exceeds the high-confidence threshold, and is wrong; otherwise it is $0$.
The
first term is large when confidence changes across policies; the second is large when the
forced-answer baseline makes a confident mistake. We use a single high-confidence cutoff of
0.8 and set $\lambda=1$ throughout.

Because the overconfident-wrong term uses correctness, $H_{\mathrm{proxy}}$ is a labeled
audit diagnostic rather than a deployment-time estimator for unlabeled inputs. Its purpose
is not to replace direct outcome diagnostics such as C0 error rate or C0 Brier risk, but to
separate two intervention-relevant signals: policy movement and overconfident-error mass.
Operationally, a larger value means that a domain shows policy movement, overconfident
failure, or both. The linear--Gaussian picture is only a guide: it illustrates why local
stability and correctness can separate, making these observable signals natural audit
quantities.

Two domains can therefore have similar average accuracy and still receive different proxy
scores. One may have relatively stable confidence and few confident mistakes, while another
may look safe on average but become risky once nearby policies or abstention are introduced.
The score is designed for this second question: not ``which domain is hardest in the
abstract,'' but ``which domain is most exposed to costly overconfidence and most likely to
benefit from intervention.''

\section{Experimental Setup}
\label{sec:setup}

We use $N=532$ short binary factual items grouped into 11 domains. This is a frozen audit
set, not a named public benchmark; the reproduction bundle includes the item strings, domain
labels, gold labels, and policy outputs. Domain sizes range from 35 to 64 items. The domains
cover medical epidemiology, social stats, geo travel, cultural industry, dev region,
entertainment event, literature media, macro index, history diplomacy, sports, and technical
standard. Every item is evaluated under the same three policies, so domain-wise comparisons
are paired throughout.
The appendix gives representative item examples, and the artifact repository
contains the frozen inputs used for all reported numbers~\cite{okutomi2026eimlartifact}.

The item file was frozen before policy evaluation. Binary gold labels were assigned
independently of model outputs, and items with missing or ambiguous written truth conditions
were excluded from the frozen audit set. Domain labels are used only for aggregation and are
not shown to the model.

\paragraph{Policies.}
C0 is a forced-answer baseline that must output ``Yes'' or ``No''. C1 allows
abstention when the item is too under-specified to answer responsibly. C2 adds an explicit
self-critical check before the final decision and may also abstain. All three policies use
the same item wording; only the decision policy changes.

\paragraph{Output-level policy model.}
All C0--C2 policy logs were collected with ``gpt-4.1-mini''. The logging script did
not set an explicit temperature parameter; exact replay may therefore depend on the API
default and hosted model snapshot. Each policy is prompted to return a decision and a verbal
$P(\mathrm{correct})$ score; the gold label is never included in the prompt. We use this
verbal confidence as a reported confidence signal, not as a calibrated probability
guaranteed by the model. We release the frozen policy-output CSV used for the output-level
analyses.

\paragraph{Confidence and policy-aware loss.}
When a policy answers, it also reports $P(\mathrm{correct}) \in [0,1]$. If it abstains, we
assign the neutral value $0.5$, which yields a fixed squared-loss penalty of $0.25$. We
evaluate policy gains with a Brier-style squared loss~\cite{brier1950,gneiting2007} on the
probability assigned to the positive ``Yes'' label. Domain-wise gain is always
measured relative to C0.

This design matters because the evaluation is intentionally intervention-aware. A policy can
help either by improving the answer itself or by refusing where the baseline tends to be
confidently wrong. The loss therefore rewards useful abstention without treating abstention
as cost-free.

For an item with binary label $y_i \in \{0,1\}$, we write
\begin{equation}
SE_{\mathrm{policy}}(i,c) = (\tilde p_{i,c}-y_i)^2,
\end{equation}
where $\tilde p_{i,c}$ is the probability assigned to the positive ``Yes'' label
under policy $c$. If the policy answers ``Yes'', then
$\tilde p_{i,c}=P(\mathrm{correct})$; if it answers ``No'', then
$\tilde p_{i,c}=1-P(\mathrm{correct})$; and if it abstains, then
$\tilde p_{i,c}=0.5$. Domain-wise gains are paired averages of the change in this loss
relative to C0.

\paragraph{Internal probe.}
To probe local internal sensitivity, we use three open-weight instruction-tuned models:
Llama-3.1-8B-Instruct, DeepSeek-R1-Distill-Llama-8B, and Qwen2.5-7B-Instruct. We add
Gaussian embedding perturbations with scale $\sigma = 0.01$ and average over 40 trials per
item. The confidently correct (CC) and overconfidently wrong (OCW) groups are inherited
from the frozen ``gpt-4.1-mini'' C0 audit logs, not recomputed from each open-weight model's
own answers. An item is CC when that C0 log is correct with reported confidence at least
0.8, and OCW when that C0 log is wrong with reported confidence at least 0.8. The threshold
is a fixed high-confidence cutoff used for this audit, not an optimized constant. Thus, the
probe asks whether audit-model high-confidence failures show greater hidden-state
sensitivity when the same item text is run through open-weight models. We then compare the
same items under a self-critical prompt that asks the model to check for missing evidence or
counter-considerations before answering. This comparison also tests whether self-critique
induces prompt-level local stabilization in hidden states, not only whether OCW items are
more fragile. Section~\ref{sec:results} also reports denser
$\sigma$ sweeps and a curated semantic-rewrite check on two models, using 12 confidently
correct and 12 overconfidently wrong items per model.

At each chosen layer, the probe measures how much the final-token hidden state moves when
the input embedding is perturbed. We use this signal only as a local diagnostic. A larger
value means that the representation moves more under the perturbation; it does not by itself
identify the source of the movement.

Concretely, let $\epsilon$ be the injected embedding-space perturbation, with norm controlled
by the noise scale $\sigma$. If $h_\ell^{\mathrm{clean}}$ and
$h_\ell^{\mathrm{noisy}}$ are the clean and perturbed final-token hidden states at layer
$\ell$, then the per-trial local sensitivity is
\begin{equation}
S_\ell(\epsilon) =
\frac{\|h_\ell^{\mathrm{noisy}}-h_\ell^{\mathrm{clean}}\|_2}{\|\epsilon\|_2}.
\end{equation}
We average this quantity over trials, then compare means across models, prompts, and the
audit-defined CC/OCW subsets.

\begin{table}[t]
  \centering
  \small
  \caption{\textbf{Compact setup summary.}}
  \label{tab:setup-summary}
  \resizebox{\linewidth}{!}{%
  \begin{tabular}{ll}
    \toprule
    Items / domains & $N=532$ / $11$ domains \\
    Policies & C0, C1, C2 \\
    High-confidence cutoff & fixed at 0.8 \\
    Abstention loss & $0.25$ \\
    Probe models & Llama-3.1 / DeepSeek-R1 / Qwen2.5 \\
    Probe noise / trials & $\sigma=0.01$, 40 trials \\
    \bottomrule
  \end{tabular}
  }
\end{table}

\section{Results}
\label{sec:results}

\subsection{Output-level audit proxy: where self-critique helps}

\begin{figure}[h]
  \centering
  \includegraphics[width=0.86\linewidth]{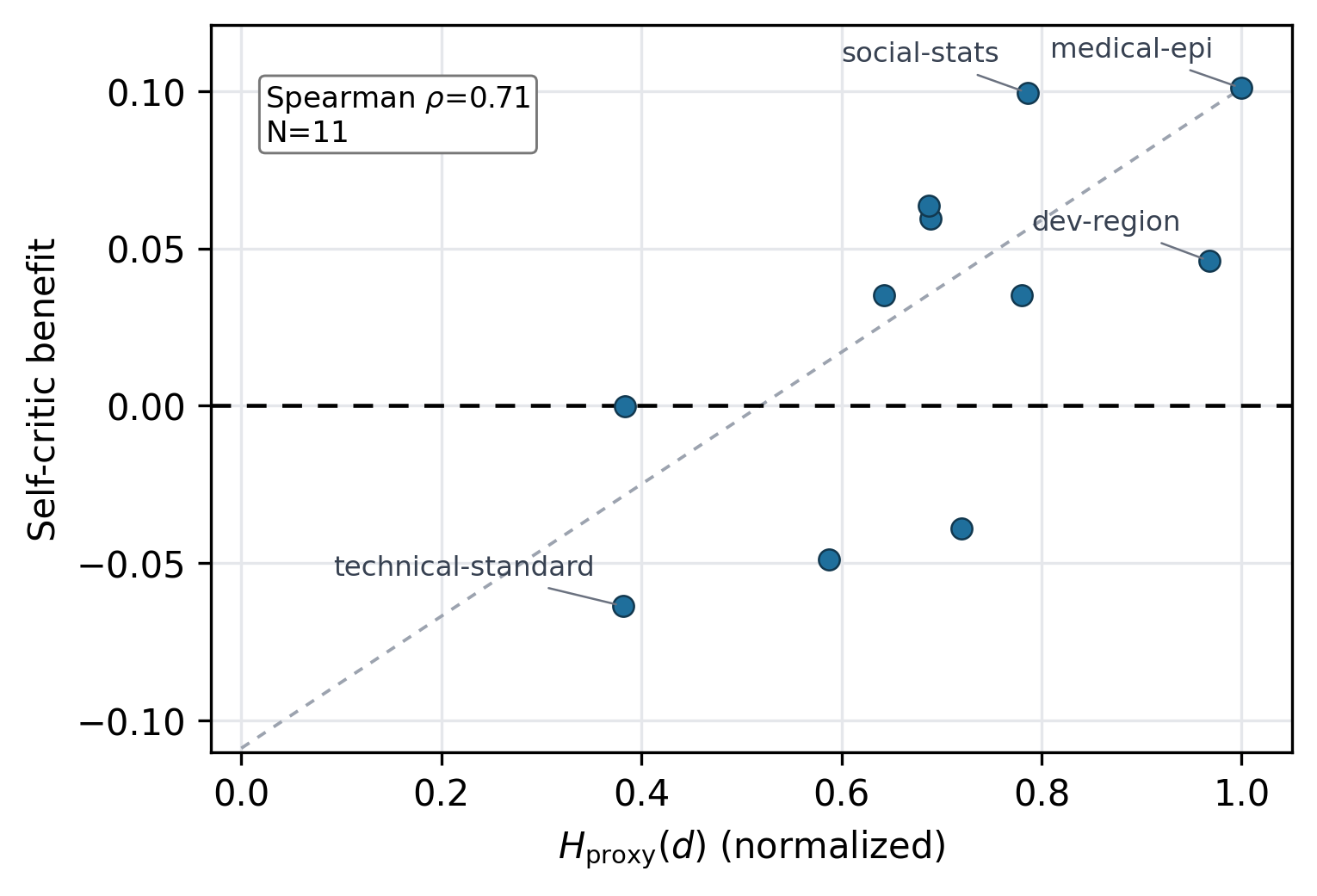}
  \caption{\textbf{$H_{\mathrm{proxy}}$ tracks where self-critique helps on the labeled
    audit set.} Each point is a domain. The dashed line is a Theil--Sen fit shown only for
    visualization.}
  \label{fig:proxy_predicts_gain}
\end{figure}

\begin{figure*}[t]
  \centering
  \includegraphics[width=0.60\textwidth]{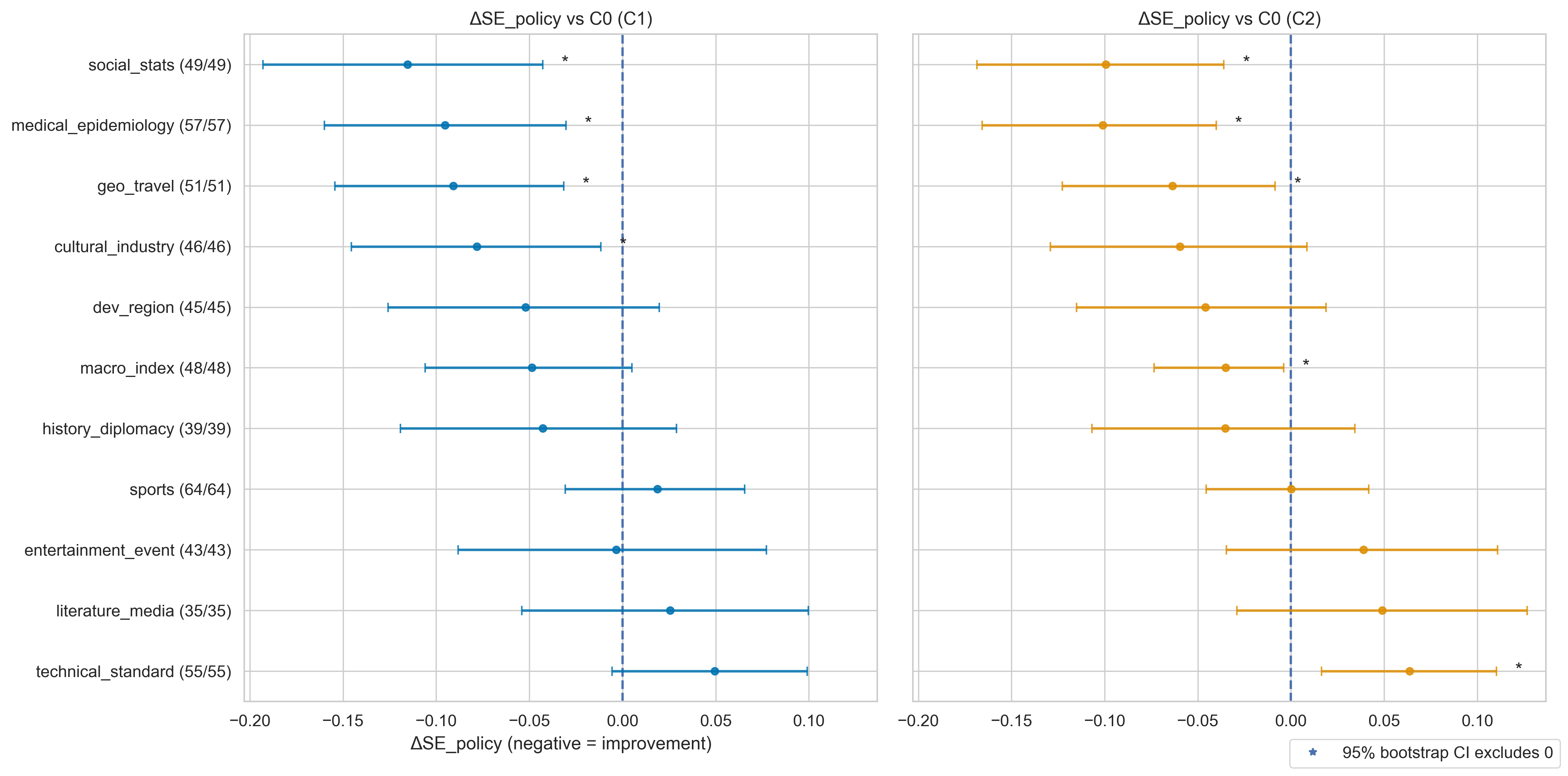}
  \caption{\textbf{Domain-wise change in policy-aware loss relative to C0.} Negative values
    indicate lower policy-aware loss than the forced-answer baseline. The domain pattern is
    mixed rather than uniform: some domains benefit substantially from abstention-aware
    self-critique, while others regress.}
  \label{fig:eiml-domain-breakdown}
\end{figure*}

Figure~\ref{fig:proxy_predicts_gain} plots normalized $H_{\mathrm{proxy}}(d)$ against the
domain-wise gain of C2 over C0. We define this gain as
$G_{C2}(d)=\overline{SE}_{C0}(d)-\overline{SE}_{C2}(d)$, so positive values indicate
improvement. In Figure~\ref{fig:eiml-domain-breakdown}, we instead plot
$\Delta SE_c(d)=\overline{SE}_{c}(d)-\overline{SE}_{C0}(d)$, so negative values indicate
improvement. The association in Figure~\ref{fig:proxy_predicts_gain} is positive: domains
with larger proxy values tend to be the domains where self-critique reduces policy-aware
loss most. The rank correlation is Spearman $\rho=0.71$ with a bootstrap 95\% interval of
$[0.12, 0.97]$.

The domain pattern is also interpretable. Medical epidemiology and social stats show the
largest gains under C2, while geo travel and cultural industry also improve. By contrast,
entertainment event and literature media do not benefit. We therefore interpret
$H_{\mathrm{proxy}}$ as a \emph{domain-ranking} signal for intervention value, not as a
causal estimate.

\begin{table}[h]
  \centering
  \scriptsize
  \setlength{\tabcolsep}{3pt}
  \caption{\textbf{Domain-ranking baselines.} Spearman correlations compare each score with
    C2 gain over C0 across 11 domains. Predictive entropy is computed from C0's reported
    confidence. Label-aware rows use gold correctness and should be read as retrospective
    audit baselines, not deployment-time scores.}
  \label{tab:uncertainty-baselines}
  \resizebox{\linewidth}{!}{\begin{tabular}{lcc}
\toprule
Domain score & Uses labels? & Spearman $\rho$ \\
\midrule
Predictive entropy (C0) & No & $-0.07$ $[-0.72, 0.73]$ \\
Confidence variation only & No & $-0.04$ $[-0.65, 0.64]$ \\
OCW rate only & Yes & $0.71$ $[0.18, 0.97]$ \\
$H_{\mathrm{proxy}}$ & Yes & $0.71$ $[0.12, 0.97]$ \\
C0 error rate & Yes & $0.85$ $[0.52, 1.00]$ \\
C0 Brier risk & Yes & $0.88$ $[0.55, 1.00]$ \\
\bottomrule
\end{tabular}
}
\end{table}

Table~\ref{tab:uncertainty-baselines} makes the baseline comparison explicit. Predictive
entropy and confidence variation alone do not rank the domains where C2 helps. Label-aware
C0 error and Brier risk rank them strongly, as expected, because they measure observed
failures directly, and in this small domain-level comparison they are stronger pure rankers
than $H_{\mathrm{proxy}}$. We therefore do not claim that $H_{\mathrm{proxy}}$ is the best
predictor of C2 gain; it is a structured diagnostic showing whether high-confidence baseline
errors coincide with policy movement, rather than a performance-optimized predictor. A full
semantic-entropy baseline would require multiple sampled answers and semantic
clustering~\cite{farquhar2024semanticentropy}; the single-answer policy logs do not contain
that distribution, so we leave that comparison to a separate multi-sample follow-up.

Table~\ref{tab:selective} shows the abstention trade-off directly. C1 is the most
conservative: it reduces the overall overconfident-wrong rate from $0.211$ to $0.028$, but
coverage falls to $0.427$. C2 keeps more coverage ($0.571$) while still lowering the
overconfident-wrong rate to $0.064$. Taken together, Figure~\ref{fig:proxy_predicts_gain}
and Table~\ref{tab:selective} support a limited practical message: the score helps describe
\emph{why} a domain may benefit from abstention-aware intervention, even when simpler
labeled diagnostics rank the domains more strongly.

The domain breakdown is also important because the gains are not uniform. Under C2, the
strongest paired improvements appear in medical epidemiology ($-0.101$), social stats
($-0.099$), and geo travel ($-0.064$), while entertainment event ($+0.039$), literature
media ($+0.049$), and technical standard ($+0.064$) become worse.

Figure~\ref{fig:eiml-domain-breakdown} makes this heterogeneity visible. C1 and C2 are not
globally better than C0 in every domain; their value lies in shifting the error profile in
places where confident mistakes are common and abstention can be used productively.

\begin{table}[h]
  \centering
  \small
  \caption{\textbf{Selective metrics under abstention-aware policies.} Coverage is answer
    rate; answer yield treats abstentions as incorrect.}
  \label{tab:selective}
  \resizebox{\linewidth}{!}{\begin{tabular}{lrrrrr}
\toprule
Condition & Coverage & Sel. Acc. & Sel. Risk & Answer Yield & OC-Wrong (overall) \\
\midrule
C0 & 1.000 & 0.639 & 0.361 & 0.639 & 0.211 \\
C1 & 0.427 & 0.744 & 0.256 & 0.318 & 0.028 \\
C2 & 0.571 & 0.674 & 0.326 & 0.385 & 0.064 \\
\bottomrule
\end{tabular}}
\end{table}

\subsection{Internal probe: prompt-induced stabilization without a clear fragility gap}

The internal probe separates a prompt-level effect from a group-level fragility test.
At the prompt level, self-critical prompting makes hidden states less locally responsive,
lowering layer-wise sensitivity $S_\ell$ across depth in all three representative models.
This supports an internal stabilization interpretation rather than a purely output-level
abstention pattern. The boundary is equally important: lower sensitivity is a stability
signal, not calibration. The reduction does not produce a clear separation between
audit-defined confidently correct (CC) and overconfidently wrong (OCW) items.

\begin{figure*}[t]
  \centering
  \includegraphics[width=\textwidth]{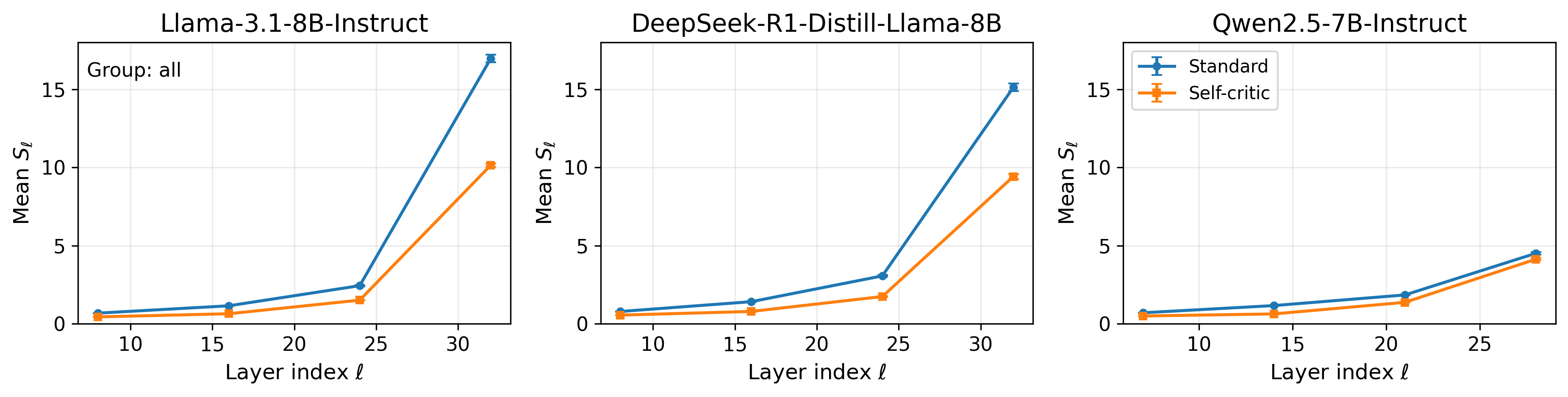}
  \caption{\textbf{Prompt-induced local stabilization.} The self-critical prompt lowers mean
    layer-wise sensitivity $S_\ell$ across depth in all three representative models; error
    bars show standard errors over items. Lower sensitivity indicates a smaller hidden-state
    response to the injected local perturbation, not evidence of calibrated correctness.}
  \label{fig:eiml-internal}
\end{figure*}

Table~\ref{tab:internal-summary} gives the corresponding final-layer numbers. In short, the
self-critical prompt lowers the overall sensitivity in all three models, while the
audit-defined OCW--CC gaps remain small at the tested scale. The bootstrap intervals include
zero, and the standardized gaps are about $0.13$ pooled standard deviations or less. We
therefore read the residual differences as weak drift, not evidence that the audit-defined
overconfident-error items are uniquely fragile.

\begin{figure*}[!b]
  \centering
  \begin{minipage}[t]{0.36\textwidth}
    \centering
    \includegraphics[width=\linewidth]{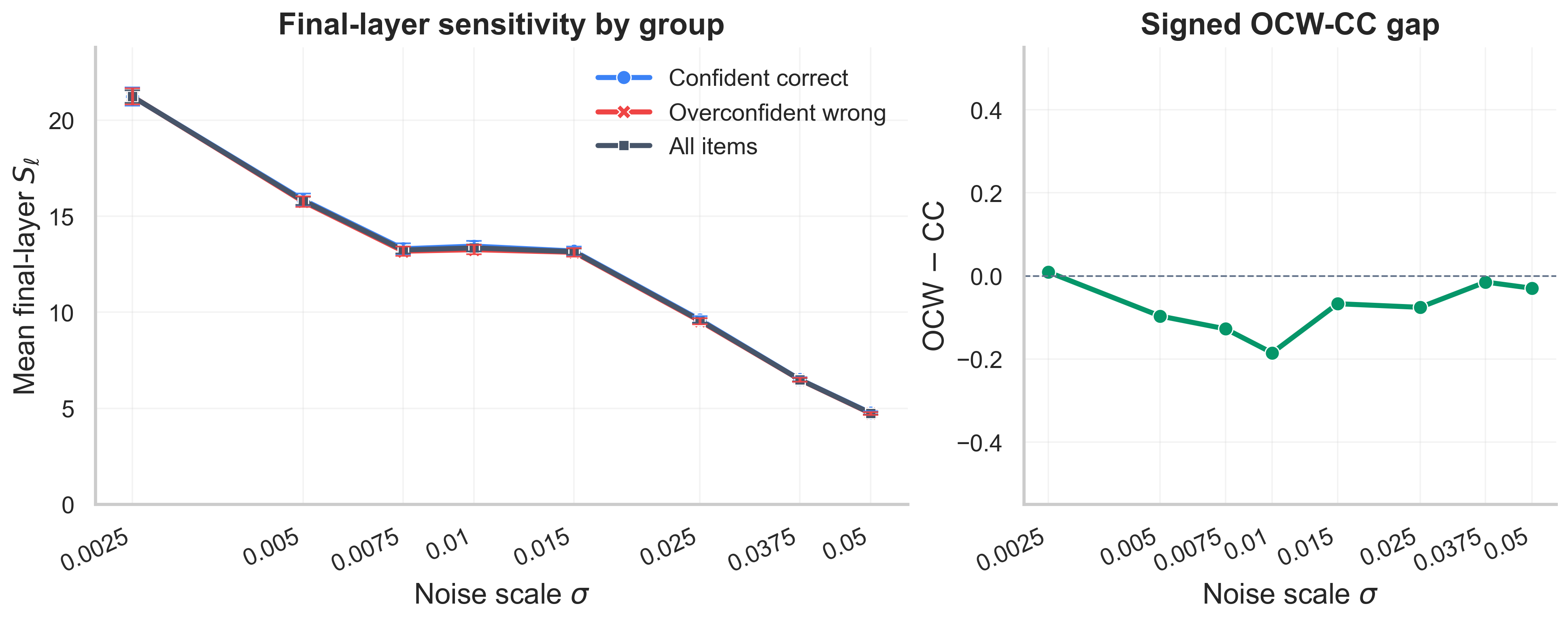}\\[0.2em]
    \includegraphics[width=\linewidth]{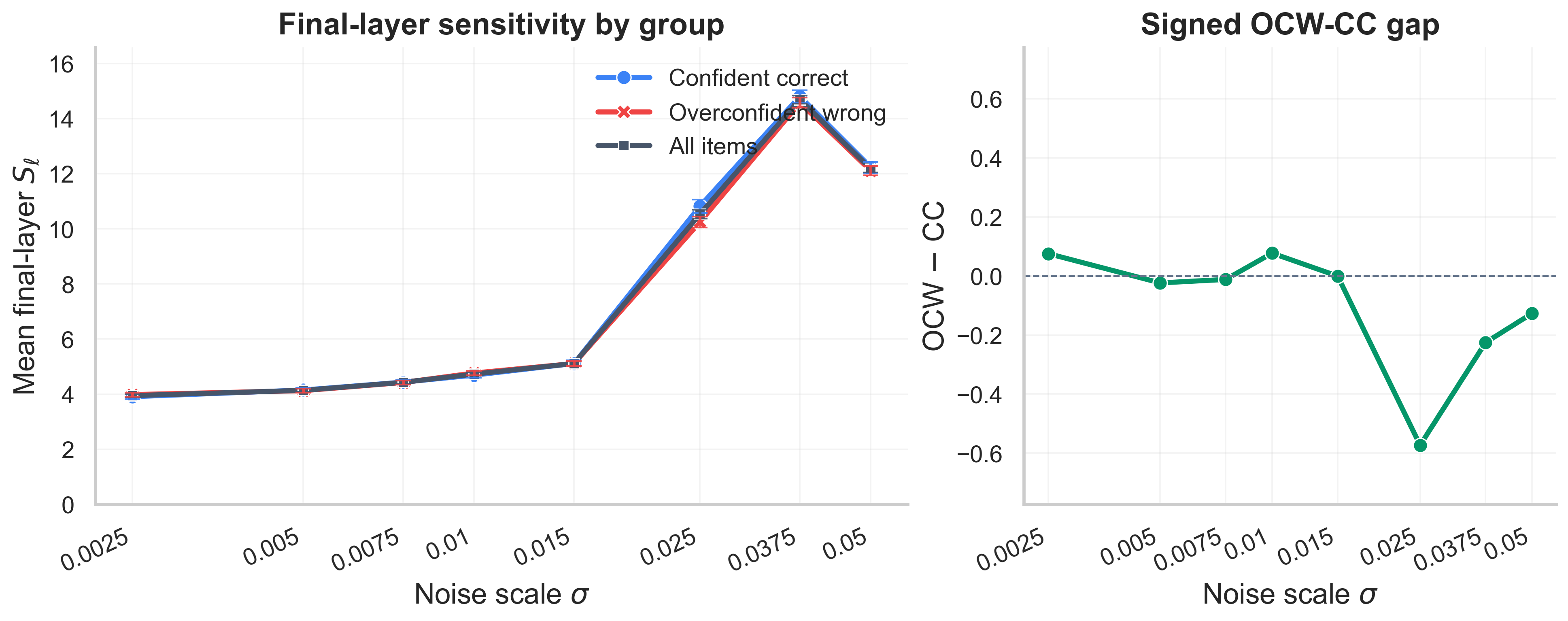}
  \end{minipage}\hfill
  \begin{minipage}[t]{0.32\textwidth}
    \centering
    \includegraphics[width=\linewidth]{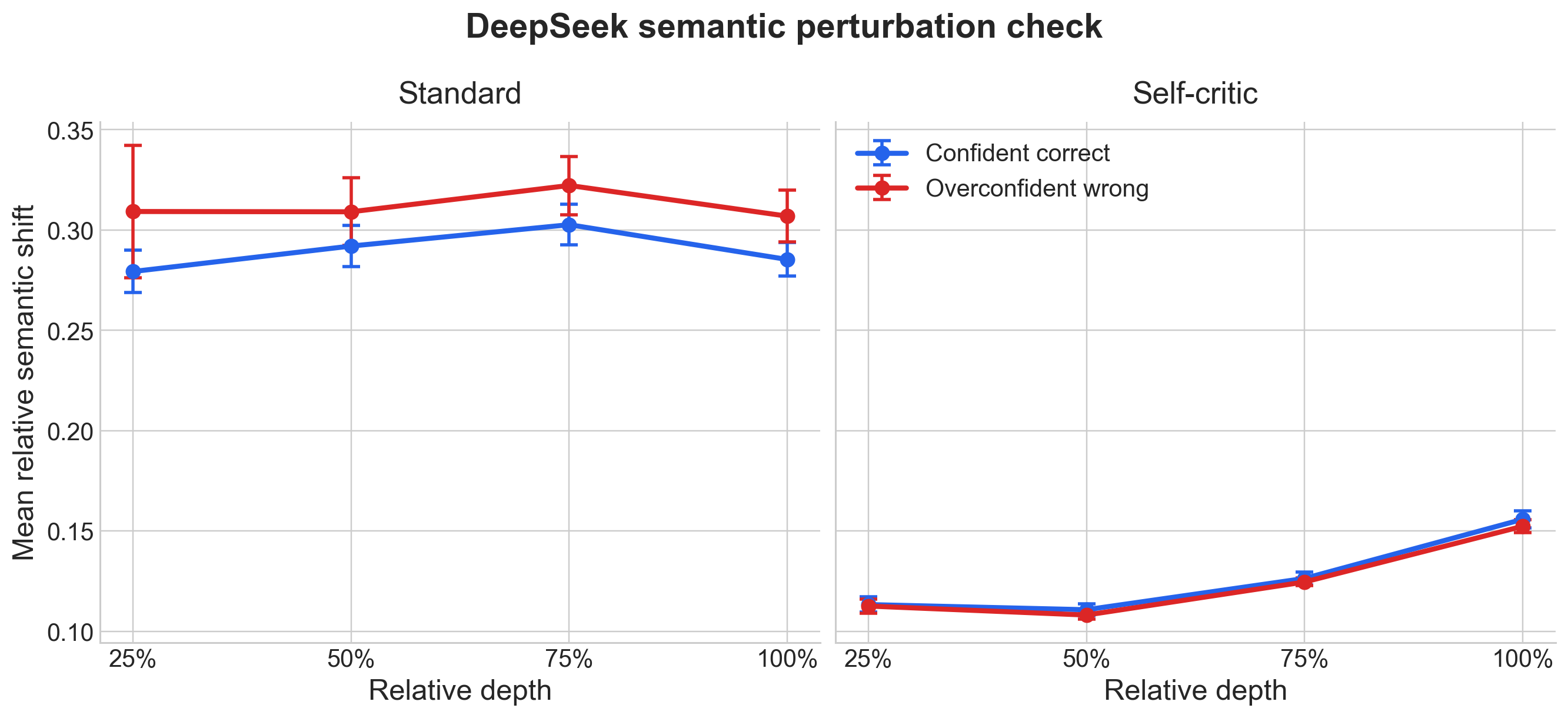}\\[0.2em]
    \includegraphics[width=\linewidth]{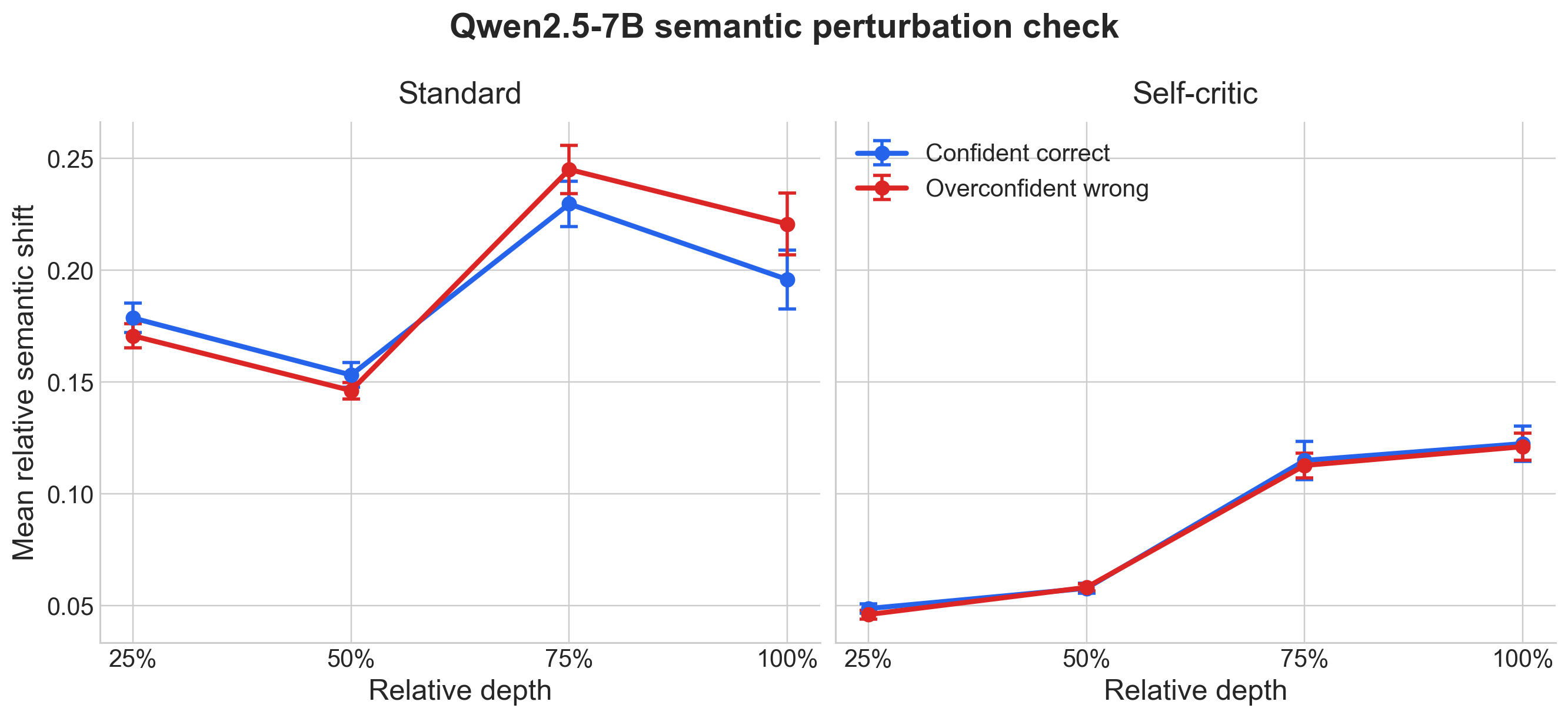}
  \end{minipage}
  \caption{\textbf{Robustness checks.} DeepSeek-R1 (top) and Qwen2.5-7B (bottom) under
    denser Gaussian sweeps (left) and semantic rewrites (right). The audit-defined OCW--CC
    gap remains small and does not steadily grow across either perturbation family.}
  \label{fig:eiml-robustness}
\end{figure*}

We also checked whether this conclusion depends on the default perturbation choice.
Figure~\ref{fig:eiml-robustness} shows denser Gaussian sweeps for DeepSeek-R1 and Qwen2.5-7B
and curated semantic rewrites for the same two models; the rewrites preserve the intended
truth condition while changing surface wording and context. These checks tell the same
story: the OCW--CC gap stays small and does not steadily grow across either perturbation
family. They are still limited, but they make the ``no clear gap'' result less likely to be
an artifact of the single default Gaussian setting.

The two result blocks answer different uncertainty questions. The output-level proxy asks
where intervention changes decision quality; it does not by itself establish stable
miscalibration. The internal probe asks a narrower follow-up question: whether items marked
as overconfidently wrong by the audit logs look uniquely fragile in open-weight models. Our
data support the intervention-ranking claim more clearly than any mechanism claim.

\begin{table}[t]
  \centering
  \small
  \setlength{\tabcolsep}{4pt}
  \caption{\textbf{Internal sensitivity summary.} ``All'' columns report final-layer mean
    sensitivity over all items under the standard and self-critical (SC) prompts. ``CC'' and
    ``OCW'' are inherited from the frozen C0 audit logs and report final-layer sensitivity
    under the standard prompt only. Gap is OCW--CC with a bootstrap 95\% CI over items;
    standardized gap is in pooled-SD units.}
  \label{tab:internal-summary}
  \resizebox{\linewidth}{!}{\begin{tabular}{lcrrrrrr}
\toprule
Model & $n_{\mathrm{CC}}/n_{\mathrm{OCW}}$ & All Std & All SC & CC & OCW & OCW--CC [95\% CI] & Std. gap \\
\midrule
Llama-3.1-8B & 40/40 & 16.97 & 10.15 & 16.82 & 17.11 & +0.30 [-0.65, +1.28] & +0.13 \\
DeepSeek-R1 & 40/40 & 15.12 & 9.42 & 14.98 & 15.26 & +0.29 [-0.70, +1.22] & +0.13 \\
Qwen2.5-7B & 40/40 & 4.51 & 4.13 & 4.51 & 4.51 & -0.00 [-0.32, +0.32] & -0.00 \\
\bottomrule
\end{tabular}
}
\end{table}

\section{Discussion and Limitations}
\label{sec:discussion}

The claim is narrow. The study uses a frozen 532-item binary factual audit set rather than a
broad public benchmark, so the rankings may not transfer to larger datasets, multi-class
tasks, or open-ended generation. $H_{\mathrm{proxy}}$ is also a labeled ranking signal, not
an absolute instability measure or a label-free deployment estimator; in this audit, the
direct labeled baselines in Table~\ref{tab:uncertainty-baselines} rank C2 gain more strongly.
Because the audit set is author-curated rather than a public benchmark with independent
multi-annotator validation, label errors or item artifacts may affect the rankings. Finally,
the bridge from stability intuition to empirical score is local. The output-level audit and
internal probe also use different model families: frozen policy logs establish the
intervention pattern, while open-weight models test whether an analogous CC--OCW fragility
gap appears where hidden states are accessible. Thus the probe does not isolate each
open-weight model's own high-confidence errors.

Thus, ``no clear CC--OCW gap'' does not mean that hidden-state instability never matters. It
means that, under the present probe, we do not see evidence for that local-fragility account.
Consequently, prompt-induced local quieting should be treated as a stability signal, not as
evidence of calibrated correctness.
Future work should test the audit score on larger public benchmarks, extend it to
multi-class and open-ended tasks, and develop label-free approximations based on answer
consistency, semantic entropy, calibrated confidence, or retrieval disagreement. It should
also compare more directly with calibration and uncertainty-estimation methods, and test
whether the CC--OCW gap changes under different confidence cutoffs, perturbation families,
model-specific CC/OCW definitions, and practical effect-size rules; the present claim is
labeled audit triage.

Stable miscalibration is also a trust risk: a system can appear robust because its answers
are locally stable, while still giving confident wrong answers in high-impact domains. The
audit should therefore be used as a periodic labeled evaluation workflow rather than as a
live safety filter: auditors can collect domain-specific labeled items, rank domains by
overconfident-error mass and policy movement, and allocate abstention, retrieval, or
human-review interventions where risk is concentrated.

\section*{Impact Statement}
This work studies high-confidence errors and abstention-aware auditing for large language
models, with the intended impact of improving reliability under uncertainty. The diagnostics
should be used for labeled auditing, risk reporting, and human review, not as a guarantee of
correctness or as a way to hide uncertainty.

\paragraph{Reproducibility.}
{\raggedright
The reproducibility artifact is archived at
\href{https://osf.io/ndz69/?view_only=0920b52c74af4ded84f0086d5d773457}{\nolinkurl{https://osf.io/ndz69/}}.
After extracting ``OSF.zip'', start with ``README.md''; the longer guide is
``paper/eiml2026/repro/README.md''. The main command is
``python scripts/reproduce\_eiml.py''. The default reproduction path uses frozen CSV inputs
and does not call external APIs.\par}

\paragraph{Disclosure.}
LLMs were used for language editing and synthetic item drafting. The author reviewed all
claims, labels, code, and results.

The views expressed are solely those of the author and do not represent any employer,
client, or affiliated organization; no employer, client, proprietary, or personal data was
used.

\section*{Appendix}
\begingroup
\footnotesize
\setlength{\parskip}{0pt}

\paragraph{Figure 4 detail.}
For the four-domain DeepSeek subset, $\sigma$ ranges from $0.0025$ to $0.05$. Across
eight scales, the signed final-layer audit-defined OCW--CC gap ranges from $-0.185$ to
$+0.009$, so this subset shows no stable positive OCW sensitivity gap.

\paragraph{Audit materials.}
The OSF repository (Okutomi, 2026) contains the audit data, prompts, examples, and
policy outputs. C0/C1/C2 are forced-answer, cautious-abstention, and self-critical; the
probe compares standard and self-critical prompts.

\endgroup

\AtBeginEnvironment{thebibliography}{\sloppy\raggedright}
\FloatBarrier
\clearpage
\bibliographystyle{icml2026}
\bibliography{refs}

@article{wen2025abstentionSurvey,
  title   = {Know Your Limits: A Survey of Abstention in Large Language Models},
  author  = {Wen, Bingbing and Yao, Jihan and Feng, Shangbin and Xu, Chenjun and Tsvetkov, Yulia and Howe, Bill and Wang, Lucy Lu},
  year    = {2025},
  journal = {Transactions of the Association for Computational Linguistics},
  volume  = {13},
  pages   = {529--556},
  url     = {https://aclanthology.org/2025.tacl-1.26/}
}

@article{tayebati2025conformalabstention,
  title   = {Learning Conformal Abstention Policies for Adaptive Risk Management in Large Language and Vision-Language Models},
  author  = {Tayebati, Sina and Kumar, Divake and Darabi, Nastaran and Jayasuriya, Dinithi and Krishnan, Ranganath and Trivedi, Amit Ranjan},
  year    = {2025},
  journal = {arXiv preprint arXiv:2502.06884},
  eprint  = {2502.06884},
  archivePrefix = {arXiv},
  primaryClass  = {cs.LG},
  url     = {https://arxiv.org/abs/2502.06884}
}

@article{an2025teachingabstain,
  title   = {Teaching {LLM}s to Abstain via Fine-Grained Semantic Confidence Reward},
  author  = {An, Hao and Xu, Yang},
  year    = {2025},
  journal = {arXiv preprint arXiv:2510.24020},
  eprint  = {2510.24020},
  archivePrefix = {arXiv},
  primaryClass  = {cs.CL},
  url     = {https://arxiv.org/abs/2510.24020}
}

@inproceedings{guo2017,
  author    = {Guo, Chuan and Pleiss, Geoff and Sun, Yu and Weinberger, Kilian Q.},
  title     = {On Calibration of Modern Neural Networks},
  booktitle = {Proceedings of the 34th International Conference on Machine Learning},
  series    = {Proceedings of Machine Learning Research},
  volume    = {70},
  pages     = {1321--1330},
  year      = {2017},
  publisher = {PMLR},
  url       = {https://proceedings.mlr.press/v70/guo17a.html}
}

@article{kadavath2022,
  title   = {Language Models (Mostly) Know What They Know},
  author  = {Kadavath, Saurav and Conerly, Thomas and Askell, Amanda and Henighan, Tom and Jones, Andy and Schiefer, Nicholas and Joseph, Nicholas and DasSarma, Nova and McCandlish, Sam and Olsson, Catherine and others},
  journal = {arXiv preprint arXiv:2207.05221},
  year    = {2022},
  eprint  = {2207.05221},
  archivePrefix = {arXiv},
  url     = {https://arxiv.org/abs/2207.05221}
}

@inproceedings{xiao2025restorecalibration,
  author    = {Xiao, Jiancong and Hou, Bojian and Wang, Zhanliang and Jin, Ruochen and Long, Qi and Su, Weijie J. and Shen, Li},
  title     = {Restoring Calibration for Aligned Large Language Models: A Calibration-Aware Fine-Tuning Approach},
  booktitle = {Proceedings of the 42nd International Conference on Machine Learning},
  series    = {Proceedings of Machine Learning Research},
  volume    = {267},
  pages     = {68364--68390},
  year      = {2025},
  publisher = {PMLR},
  url       = {https://proceedings.mlr.press/v267/xiao25b.html}
}

@article{wang2026faithfulconfidence,
  title   = {Are {LLM} Decisions Faithful to Verbal Confidence?},
  author  = {Wang, Jiawei and Zhou, Yanfei and Devic, Siddartha and Fu, Deqing},
  year    = {2026},
  journal = {arXiv preprint arXiv:2601.07767},
  eprint  = {2601.07767},
  archivePrefix = {arXiv},
  primaryClass  = {cs.AI},
  url     = {https://arxiv.org/abs/2601.07767}
}

@inproceedings{ji2024_internal_states_hallu_risk,
  author    = {Ji, Ziwei and Chen, Delong and Ishii, Etsuko and Cahyawijaya, Samuel and Bang, Yejin and Wilie, Bryan and Fung, Pascale},
  title     = {{LLM} Internal States Reveal Hallucination Risk Faced With a Query},
  booktitle = {Proceedings of the 7th BlackboxNLP Workshop: Analyzing and Interpreting Neural Networks for NLP},
  year      = {2024},
  address   = {Miami, Florida, US},
  publisher = {Association for Computational Linguistics},
  pages     = {88--104},
  doi       = {10.18653/v1/2024.blackboxnlp-1.6},
  url       = {https://aclanthology.org/2024.blackboxnlp-1.6/}
}

@inproceedings{manakul2023selfcheckgpt,
  author    = {Manakul, Potsawee and Liusie, Adian and Gales, Mark},
  title     = {{SelfCheckGPT}: Zero-Resource Black-Box Hallucination Detection for Generative Large Language Models},
  booktitle = {Proceedings of the 2023 Conference on Empirical Methods in Natural Language Processing},
  year      = {2023},
  address   = {Singapore},
  publisher = {Association for Computational Linguistics},
  pages     = {9004--9017},
  doi       = {10.18653/v1/2023.emnlp-main.557},
  url       = {https://aclanthology.org/2023.emnlp-main.557/}
}

@article{joo2025consistency,
  title   = {Black-Box Hallucination Detection via Consistency Under the Uncertain Expression},
  author  = {Joo, Seongho and Min, Kyungmin and Koo, Jahyun and Jung, Kyomin},
  year    = {2025},
  eprint  = {2509.21999},
  archivePrefix = {arXiv},
  primaryClass = {cs.CL},
  journal = {arXiv preprint arXiv:2509.21999}
}

@article{ji2023survey,
  author  = {Ji, Ziwei and Lee, Nayeon and Frieske, Rita and Yu, Tiezheng and Su, Dan and Xu, Yan and Ishii, Etsuko and Bang, Ye Jin and Madotto, Andrea and Fung, Pascale},
  title   = {Survey of Hallucination in Natural Language Generation},
  journal = {ACM Computing Surveys},
  volume  = {55},
  number  = {12},
  pages   = {1--38},
  year    = {2023},
  doi     = {10.1145/3571730},
  url     = {https://doi.org/10.1145/3571730}
}

@article{kalman1960,
  author  = {Kalman, R. E.},
  title   = {A New Approach to Linear Filtering and Prediction Problems},
  journal = {Journal of Basic Engineering},
  year    = {1960},
  volume  = {82},
  number  = {1},
  pages   = {35--45},
  doi     = {10.1115/1.3662552}
}

@article{brier1950,
  author  = {Brier, Glenn W.},
  title   = {Verification of Forecasts Expressed in Terms of Probability},
  journal = {Monthly Weather Review},
  volume  = {78},
  number  = {1},
  pages   = {1--3},
  year    = {1950},
  doi     = {10.1175/1520-0493(1950)078<0001:VOFEIT>2.0.CO;2}
}

@article{gneiting2007,
  author  = {Gneiting, Tilmann and Raftery, Adrian E.},
  title   = {Strictly Proper Scoring Rules, Prediction, and Estimation},
  journal = {Journal of the American Statistical Association},
  volume  = {102},
  number  = {477},
  pages   = {359--378},
  year    = {2007},
  doi     = {10.1198/016214506000001437}
}

@article{farquhar2024semanticentropy,
  title   = {Detecting hallucinations in large language models using semantic entropy},
  author  = {Farquhar, Sebastian and Kossen, Jannik and Kuhn, Lorenz and Gal, Yarin},
  journal = {Nature},
  volume  = {630},
  pages   = {625--630},
  year    = {2024},
  doi     = {10.1038/s41586-024-07421-0},
  url     = {https://www.nature.com/articles/s41586-024-07421-0}
}

@misc{okutomi2026eimlartifact,
  author       = {Okutomi, Akira},
  title        = {{OSF} Artifact Repository for Stable Miscalibration in Large Language Models},
  year         = {2026},
  howpublished = {Available at \href{https://osf.io/ndz69/?view_only=0920b52c74af4ded84f0086d5d773457}{\nolinkurl{https://osf.io/ndz69/}}},
  note         = {Reproducibility artifact}
}

\end{document}